 \documentclass[pmlr,twocolumn,10pt]{jmlr} % W&CP article

\usepackage{booktabs}
\usepackage{siunitx}

\usepackage[switch]{lineno}

\usepackage{changepage}
\usepackage{multirow}
\usepackage{graphicx}
\usepackage[font=small,skip=0pt]{caption}
\theorembodyfont{\upshape}
\theoremheaderfont{\scshape}
\theorempostheader{:}
\theoremsep{\newline}

\jmlrvolume{259}
\jmlryear{2024}
\jmlrsubmitted{LEAVE UNSET}
\jmlrpublished{LEAVE UNSET}
\jmlrworkshop{Machine Learning for Health (ML4H) 2024} % W&CP title

 \title[Self-supervised nursing note summarization]{Query-Guided Self-Supervised Summarization of Nursing Notes}

\author{\Name{Ya Gao~\textsuperscript{1}}
\Email{ya.gao@aalto.fi}\\
\Name{Hans Moen~\textsuperscript{1}}
\Email{hans.moen@aalto.fi}\\
\Name{Saila Koivusalo~\textsuperscript{2}}
\Email{saila.koivusalo@hus.fi}\\
\Name{Miika Koskinen~\textsuperscript{2}}
\Email{miika.koskinen@hus.fi}\\
\Name{Pekka Marttinen~\textsuperscript{1}}
\Email{pekka.marttinen@aalto.fi}\\
\addr \textsuperscript{1} Aalto University, Finland\\
\addr \textsuperscript{2} Helsinki University Hospital, Finland\\
}

\begin{document}

\maketitle
\begin{abstract}
Nursing notes, an important part of Electronic Health Records (EHRs), track a patient's health during a care episode. Summarizing key information in nursing notes can help clinicians quickly understand patients' conditions. However, existing summarization methods in the clinical setting, especially abstractive methods, have overlooked nursing notes and require reference summaries for training. We introduce QGSumm, a novel query-guided self-supervised domain adaptation approach for abstractive nursing note summarization. The method uses patient-related clinical queries for guidance, and hence does not need reference summaries for training. Through automatic experiments and manual evaluation by an expert clinician, we study our approach and other state-of-the-art Large Language Models (LLMs) for nursing note summarization. Our experiments show: 1) GPT-4 is competitive in maintaining information in the original nursing notes, 2) QGSumm can generate high-quality summaries with a good balance between recall of the original content and hallucination rate lower than other top methods. Ultimately, our work offers a new perspective on conditional text summarization, tailored to clinical applications.
\end{abstract}

\section{Introduction}
\label{sec:intro}
Nursing notes are important for clinicians to track a patient's health status and administered treatments during hospitalization \citep{tornvall2008nursing}. 
However, a care episode may result in a large number of nursing notes, especially for patients suffering from complex health problems \citep{hall2004information}. 
Furthermore, the condensed nature of nursing notes makes them time-consuming for clinicians to understand~\citep{clarke2013information}.
Text summarization methods from Natural Language Processing (NLP) help distill the content of clinical notes \citep{wang2021systematic}, making crucial information more accessible and less time-consuming to review. This can be particularly beneficial for clinicians, and for conducting retrospective studies.
These summarization techniques are divided into extractive~\citep{Pivovarov2015AutomatedMF, moen2016comparison,Tang2019ProgressNC} and abstractive~\citep{zhang2020optimizing, liu2022retrieve, searle2023discharge}.
Extractive methods retain the original sentences and/or keyphrases but may lack coherence or fluency.
On the other hand, abstractive methods produce smoother summaries but typically require explicit supervision, i.e., a reference summary as the ground truth, which are time-consuming to produce\citep{o2009physicians}.

\begin{figure*}[t]
\floatconts
{fig:example}
{\caption{From a patient's admission to discharge, multiple nursing notes may be generated. As shown in one artificial nursing note example, the notes could be poorly organized and lack clarity.}}
{\includegraphics[width=0.8\linewidth]{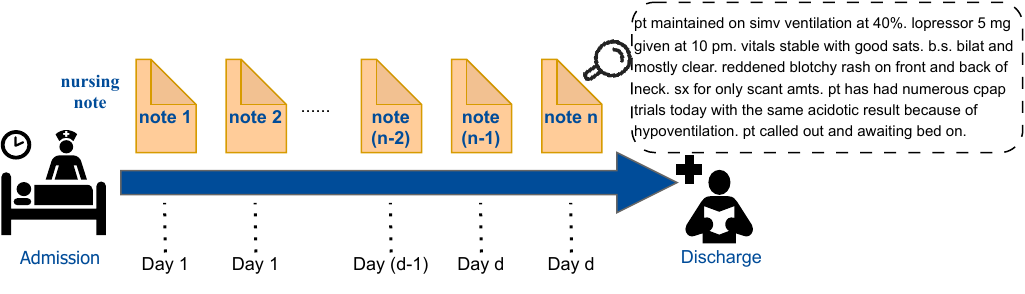}}
\end{figure*} 

Self-supervised abstractive text summarization \citep{chu2019meansum, elsahar2021self} bypasses the problem of lacking ground-truth summaries. Previous self-supervised methods (e.g., \cite{liu2021self}) reduce the semantic distance between a summary and the original text. However, unconditionally making the semantic representations similar lacks control of the generated summaries and may result in missing relevant information, which can be important especially in clinical applications.
Furthermore, the content of nursing notes may be unclear, poorly organized and contain non-standard abbreviations (Figure~\ref{fig:example}). Hence, in the absence of supervised learning signals, guiding the model to generate a good summary becomes challenging. 

In this paper, we propose a query-guided self-supervised domain adaptation framework for nursing note summarization, named QGSumm. A good summary of a clinical note is centered on the patient's condition.

We achieve this by designing queries that concern the patient's condition, and we train the model to produce summaries such that answers to the queries based on the summaries are similar to the answers based on the original notes.  
This guidance makes our method highly suitable for the clinical field since the resulting summaries are centered on the information nurses and clinicians are most concerned about.
To the best of our knowledge, our study is the first on abstractive summarization of nursing notes and on employing self-supervised learning for clinical note summarization. 

Our primary contributions are: \textbf{(1)} The study focuses on nursing notes that play a critical role in clinical settings, filling a gap in previous research by introducing a method for abstractive nursing note summarization.
Our method's ability to work without reference summaries highlights its practical applicability, while also being relatively light-weight and suitable for deployment in a secure environment with limited computing resources.
\textbf{(2)} We propose a novel self-supervised domain adaptation framework. By leveraging patient-related queries,  we guide the model to generate nursing note summaries that prioritize specific content, i.e., the patients' conditions and health status, without the need for manually written reference summaries as ground truth.
\textbf{(3)} We conduct a comprehensive automated empirical study and a manual evaluation by an expert clinician, including state-of-the-art Large Language Models (LLMs), which have not been previously investigated for nursing note summarization. The experiments demonstrate especially our method's and GPT-4's ability to perform well in this task.

\begin{figure*}[t]
   \floatconts 
   {fig:framework}
   % {\caption {The overall architecture. While fine-tuning the encoder (ENC) and decoder (DEC), $\operatorname{DEC^{rec}}$ and the query responder network $\operatorname{R}$ are frozen. Raw patient information is processed by ENC into the embeddings $\mathbf{H}^{PA}$, but this is omitted here for clarity (see Fig. \ref{fig:augmentation} for details).}}
      {\caption {\textbf{Stage 1:} Fine-tune the encoder ($\operatorname{ENC_{0}}$) by note reconstruction, where $\operatorname{DEC_{0}}$ is frozen; \textbf{Stage 2:} Fine-tune the encoder (ENC) and decoder (DEC) in the self-supervised manner, where the query responder $\operatorname{R}$ is frozen. Raw patient information is processed by ENC into the embeddings $\mathbf{H}^{PA}$, as shown in Fig. \ref{fig:augmentation}}}
   {\includegraphics[width=0.9\linewidth]{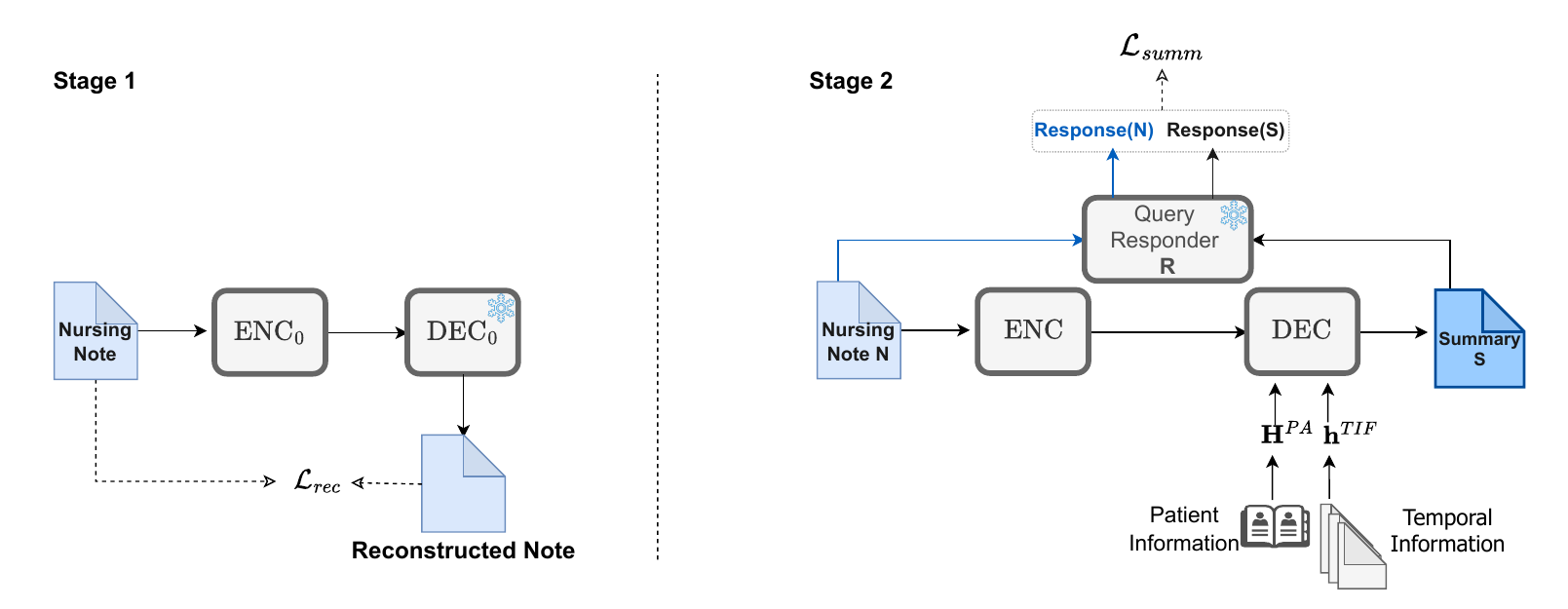}}
\end{figure*} 

\section{Related Work}
\label{sec:related_work}
% \subsection*{Key Information Extraction and Summarization from Clinical Notes}
% \subsection*{Clinical Notes Summarization}
\paragraph{Clinical Notes Summarization.}
Extractive summarization methods can preserve faithfulness but may have difficulties in maintaining coherence, and include earlier semantic similarity-based techniques \citep{Pivovarov2015AutomatedMF,moen2016comparison}, as well as more recent attention-based methods~\citep{Tang2019ProgressNC, reunamo2022text, kanwal2022attention}. %to determine key information in clinical text with an emphasis on explainability \citep{Tang2019ProgressNC, reunamo2022text, kanwal2022attention}.
%Recently, there has been a notable increase in research on abstractive clinical text summarization. 
On the other hand, abstractive clinical notes summarization has been applied on discharge summaries \citep{shing2021towards,adams2022learning, searle2023discharge}, radiology reports \citep{zhang2020optimizing,van2023radadapt,nishio2024fully}, doctor-patient conversations \citep{zhang2021leveraging, krishna2021generating, abacha2023empirical} and problem lists \citep{gao2022summarizing, gao2023overview}. However, unlike ours, these works depend on data annotation or reference summaries. LLMs demonstrate a remarkable capability in clinical text understanding. %, leading to interest in investigating their performance in clinical text summarization. 
\cite{van2024adapted} extensively analyze the clinical text summarization performance of various LLMs with in-context learning \citep{lampinen2022can} and QLoRA \citep{dettmers2024qlora} adaptation.
%, comparing the performance of LLMs with medical experts.%, providing insights into the strengths and limitations of LLMs.

% \subsection*{Unsupervised and Self-Supervised Abstractive Text Summarization}
\paragraph{Unsupervised and Self-Supervised Abstractive Text Summarization.}
The scarcity of annotated text has spurred interest in unsupervised and self-supervised text summarization.
Previous works like source document reconstruction assume that a good summary can reconstruct the source document~\citep{chu2019meansum}, but such summaries may lack specific information due to limited control.
Two-step approaches extract important information first and then perform abstractive summarization based on this extraction \citep{zhong2022unsupervised, ke2022consistsum, liu2022learning}.
However, the quality of such summaries relies on the extraction effectiveness, and developing a reliable extractor can be costly.
A contrastive learning strategy, proposed by \cite{zhuang2022learning}, aims to maximize similarity between generated summaries and source documents while minimizing it between summaries and edited documents. 
\cite{hosking2023attributable} suggest an attributable opinion summarization system that encodes sentences as paths through a hierarchical discrete latent space, identifying common subpaths for an entity to generate the summary.
\cite{jin2024self} propose using a review from a review set as the hypothetic summary to carry out self-supervised summarization of product reviews. 

\section{Methods}
\label{sec:method}
Next, we introduce QGSumm, a novel framework for summarizing clinical notes, with a focus on capturing important patient-centered information in a self-supervised fashion. 
We propose a self-supervised domain adaptation strategy applied on the base model in Section~\ref{sec:base_model}. 
This strategy positive-contrastively learns from the original nursing notes, providing the summaries with an ability comparable to the original notes to resolve patient-related queries (Section~\ref{sec:tr_objctive}).
% Moreover, we focus the model on the patient's meta information and temporal aspects by two augmentation blocks, detailed in Section \ref{sec:augmentation_block}.
Using two augmentation blocks (Section~\ref{sec:augmentation_block}), the model leverages patient metadata and temporal aspects.
% Our model summarizes one nursing note at a time, taking into account its context.
Assume a patient $PT$ has a sequence of nursing notes $N=\{N_1, N_2,\ldots, N_m\}$ sorted by time. 
Our objective is to obtain a summary $S_i$ for note $N_i$ from the distribution $P(S_i|N_i, {PA},\{N_1,\ldots,N_{i-1}\},U)$, which is conditioned on the patient's metadata ${PA}$, information in the previous notes $\{N_1,\ldots,N_{i-1}\}$, and the query $U$ which helps guide the generation.

\subsection{Base Model}
\label{sec:base_model}
\begin{figure*}[t]
\floatconts
      {fig:augmentation}
      {\caption{The proposed Temporal Information Fusion(TIF) block and the Patient Information Augmentation (PIA) block. The figure shows the process of deriving $\mathbf{H}^{dec}$ for generating the $(j+1)$th token in the summary.}}
      {\includegraphics[width=0.8\linewidth]{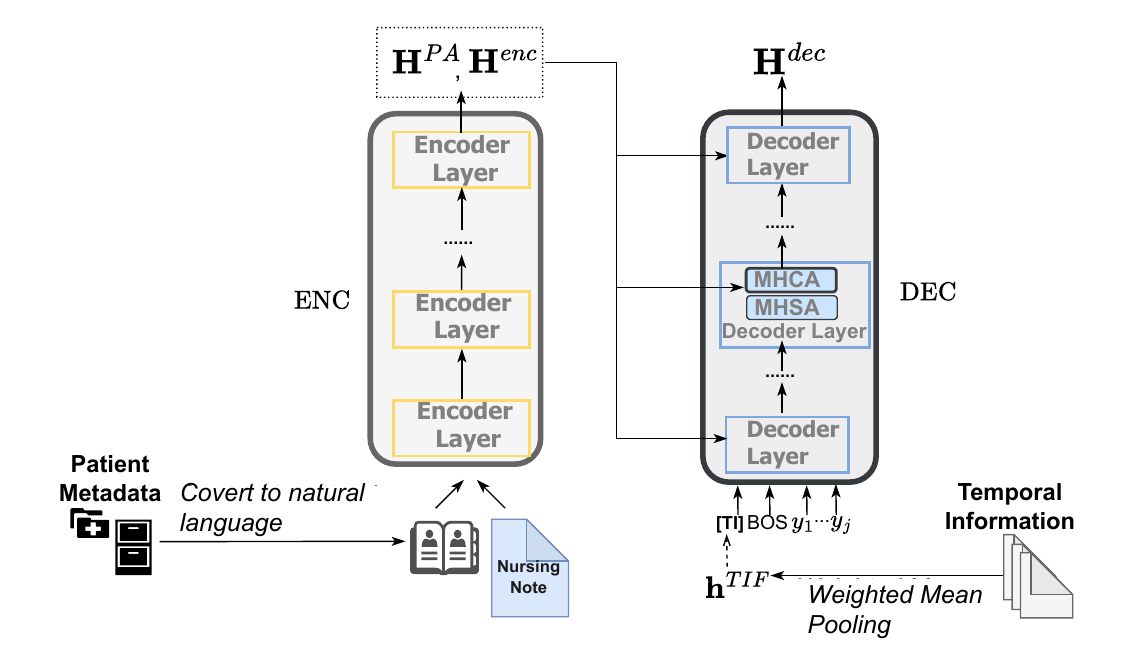}}
\end{figure*}

The backbone of our framework is an off-the-shelf transformer-based language model with an encoder-decoder structure. We leverage its checkpoint $\operatorname{M_0}$, fine-tuned for text summarization, as the \textbf{base model}, utilizing the pre-trained resources without training from scratch.
However, $\operatorname{M_0}$ exhibits limited clinical text comprehension.
Hence, in \textbf{Stage 1} we improve the clinical text understanding of $\operatorname{M_0}$'s encoder, $\operatorname{ENC_0}$, see Figure \ref{fig:framework}.
Specifically, we employ $\operatorname{ENC_0}$ from $\operatorname{M_0}$ and the frozen decoder $\operatorname{DEC_{0}}$ from the original pre-trained model, training $\operatorname{ENC_0}$ by reconstructing $N_i$.
This yields $\operatorname{M}$ with an improved encoder $\operatorname{ENC}$ which better grasps the complex semantics of nursing notes.

\subsection{Training Objective}
\label{sec:tr_objctive}
Since there is no ground truth summary available, 
in \textbf{Stage 2}, we adopt a self-supervised strategy for model $\operatorname{M}$ to generate high-quality, patient-centered summaries that can respond to patient-related queries effectively (Figure~\ref{fig:framework}).
We introduce a separate model $\operatorname{R}$, a \textbf{query responder}, trained to handle patient-related queries such as predicting readmission risk using data available in the patient database. For example, given the note $N_i$ or its summary $S_i$ as input to $\operatorname{R}$, the output will be the response to the corresponding query, like the readmission probability.
As the training objective for $\operatorname{M}$ we minimize the discrepancy, calculated as the cross-entropy loss $\mathcal{L}_{CE}$, between the two responses based on either $N_i$ or $S_i$ respectively.
This ensures that when responding to a certain patient-related query, using the summary will produce a response similar to that obtained using the nursing note, without ever showing the correct query answers to model $\operatorname{M}$.
To prevent $\operatorname{M}$ from generating summaries that are too verbose or direct ``copy-paste'' from the original notes, we introduce a length penalty term. 
The final loss function for a batch of nursing notes is: 
\begin{align}
\label{eq:training_obj}
    \mathcal{L}_{summ} &= \frac{1}{K}\sum_{i=1}^K\mathcal{L}_{CE}(\operatorname{R}(N_i),\operatorname{R}(S_i)) \nonumber\\
    &\times (1+\lambda_{1}e^{(\alpha-0.5)}),
\end{align}

where
\begin{equation}
    \alpha = \frac{\sum_{i=1}^K\operatorname{Len}(S_i)}{\sum_{i=1}^K\operatorname{Len}(N_i)},
\end{equation}
$K$ is the batch size, and $\operatorname{Len}$ denotes the number of tokens in the document. 
The hyperparameter $\lambda_1 \in [0,1]$ regulates the extent of the penalty, and is selected using the validation set, see details in~\hyperlink{sec:length_coef_effect}{C.2}.
%The value selection of it is shown in \hyperlink{sec:length_coef_effect}{C.2}.
%
Nondifferentiability from $\operatorname{M}$ and $\operatorname{R}$ is resolved using the straight-through Gumbel-softmax trick \citep{bengio2013estimating,jang2017categorical}.

\subsection{Augmentation Blocks for the Context of the Patient}
\label{sec:augmentation_block}

\paragraph{Temporal Information Fusion (TIF).}  
A patient typically has sequential nursing notes to document the evolution of their condition. 
Therefore, the context provided by the prior notes helps understanding the patient's current status. We regard this as \textit{temporal information} and include it during summarization to help the model to understand the patient's condition.
For $N_i$, the embeddings of its previous notes are represented by the embeddings of their respective first tokens, which are special tokens indicating the start of each note. These embeddings are obtained at the last hidden state in the $\operatorname{ENC}$, denoted as $\{\mathbf{h_1}, \mathbf{h_2},\ldots, \mathbf{h_{i-1}}\}$, where $\mathbf{h_i}\in\mathbb{R}^{d}$ and $d$ is the hidden state dimension. 
We aggregate past note representations via weighted mean pooling, emphasizing the most recent notes. In practice, initial weights $\beta_{j}, j=1,\ldots,i-1$ for each past note $N_i$ are based on the position in the sequence: $\beta_1=1, \beta_2=2, \ldots, \beta_{i-1}=i-1$. We use the normalized weights $\beta^{'}_{j}=\frac{\beta_j}{\beta_1+\beta_2+...+\beta_{i-1}}$ to perform the weighted mean pooling:
\begin{equation}
    \mathbf{h}^{TIF}=\operatorname{MeanPooling}(\beta_1^{'}\mathbf{h}_1, \beta_2^{'}\mathbf{h}_2,..., \beta_{i-1}^{'}\mathbf{h}_{i-1}),
\end{equation}
where $\mathbf{h}^{TIF}\in\mathbb{R}^{d}$ represents the information fusion of the past notes. 
As shown in Figure \ref{fig:augmentation}, we prepend a special token \texttt{[TI]} at the beginning of the decoder input, representing temporal information with embedding $\mathbf{h}^{TIF}$. 
Consequently, the initial input to the decoder consists of \texttt{[[TI], [BOS]]}, where \texttt{[BOS]} indicating the start of generation. 
\texttt{[TI]} is replaced with the padding token \texttt{[PAD]} for nursing notes without past notes.

The model generates tokens auto-regressively, appending produced tokens to the decoder input when generating subsequent tokens. Thus, \texttt{[TI]} consistently prompts the model to focus on the patient's past context throughout the summarization process.

\paragraph{Patient Information Augmentation (PIA).}
We aim to obtain summaries focusing on the patient's condition by incorporating patient-level information into the model via cross-attention, facilitating the interaction of information on different levels.
A patient's metadata PA typically comprises basic information, including age, gender, existing diagnoses, and performed procedures.
We convert this metadata into patient information in natural language (one example in \hyperlink{sec:metadata}{A.1}), then encode it using $\operatorname{ENC}$ to derive patient embedding $\mathbf{H}^{PA} \in \mathbb{R}^{z \times d}$, where $z$ represents the number of tokens in patient information. The source note is also encoded into embedding $\mathbf{H}^{enc} \in \mathbb{R}^{n \times d}$, where $n$ denotes the number of tokens in the note given as input.
On the decoder $\operatorname{DEC}$ side, let us assume the tokens input to the decoder at the current timestep are \texttt{[[TI], [BOS], $y_1$,\ldots,$y_{j}$]}. Consequently, the hidden representation passed to the $l$th decoder layer is $\mathbf{H}_{l}^{dec} \in \mathbb{R}^{{(j+2)} \times d}$. 
$\mathbf{H}_{l}^{dec}$ is updated using Multi-Head Self-Attention ($\operatorname{MHSA}$) and Multi-Head Cross-Attention ($\operatorname{MHCA}$)~\citep{vaswani2017attention} involving $\mathbf{H}^{enc}$ and $\mathbf{H}^{PA}$.
This facilitates the fusion of patient and note-level information:
\begin{align}
\label{eq:PIA}
    \mathbf{H}_{l+1}^{dec} &= \operatorname{MHCA}(\mathbf{H}^{enc}, \operatorname{MHSA}(\mathbf{H}_{l}^{dec})) \nonumber\\&+ \lambda_2\times\operatorname{MHCA}(\mathbf{H}^{PA}, \operatorname{MHSA}(\mathbf{H}_{l}^{dec})),
\end{align}
$\lambda_2 \in [0,1]$ is a hyperparameter to control the importance of patient meta information, selected using the validation set, see details in \hyperlink{sec:importance_effect}{C.3}.
$\mathbf{H}_{l+1}^{dec}$ is the input to the next decoder layer, or if the $l$th layer is the final layer, it is the input to the language modeling head.

With these two augmentation blocks, we can obtain the final decoder state $\mathbf{H}^{dec} \in \mathbb{R}^{(j+2) \times d}$, which is input to the language modeling head for generating the $(j+1)$th token in the summary.
The computation is shown in \hyperlink{sec:next_token}{A.5}

\section{Experimental setup} 
\subsection{Data}
We use MIMIC-III \citep{johnson2016mimic}, an EHR database with clinical notes organized by admission. Notes are treated independently per admission due to the discontinuity between admissions of the same patient. 
We focus on the nursing notes within the clinical notes.
After preprocessing (details in \hyperlink{sec:data}{A.2}), the numbers of notes in the training, validation and test sets are 149015, 10001, 3079 and the corresponding numbers of admissions are 13893, 1000, 1156.

\subsection{Types of Queries}
\label{sec:queries}
Two principles guide the query selection:
(a) The query should be related to the patient and learnable by the query responder $\operatorname{R}$; 
(b) Training data for $\operatorname{R}$ should be easily available. 
We propose two queries: \textbf{Readmission Prediction}, predicting if a patient will be readmitted within 30 days post-discharge, and \textbf{Phenotype Classification}, identifying which of 25 phenotypes the patient has~\citep{harutyunyan2019multitask}. In the experiments, we combine these queries.

The query responder $\operatorname{R}$ is a classifier that predicts an answer to the query. Part of the training data is used to train $\operatorname{R}$. When using $\operatorname{R}$, we input the summary or the original note to predict the classification probabilities, and minimize the discrepancy between these predictions (see Section \ref{sec:tr_objctive}). As a baseline we include a query by minimizing the cosine similarity between the note and its summary. More details on queries and their combination are in \hyperlink{sec:uq}{A.3}.

\subsection{Detailed Settings}
We use BART-Large-CNN\footnote{\url{https://huggingface.co/facebook/bart-large-cnn}} as the \textbf{base model}, a BART model \citep{lewis2020bart} fine-tuned for text summarization. For the \textbf{query responder}, we use Clinical-Longformer \citep{li2022clinical}, as it can handle long contexts, and fine-tune it on the selected queries. Hyperparameters for the models are specified in \hyperlink{sec:hp}{A.4}.

Since our method is designed for scenarios where reference summaries are unavailable, we compare our method with \textbf{baselines} in zero-shot settings: BART-Large-CNN (\textit{BART}) and Pegasus \citep{zhang2020pegasus}; and in 1-shot in-context learning prompting using GPT-4 \citep{openai2023gpt} and BioMistral-7B \citep{labrak2024biomistral} (\textit{BioMistral}). Consistent with QGSumm, we fine-tune the encoder of BART and Pegasus to reconstruct notes, enhancing their performance. We consider BART as the most relevant baseline for our method as it is the base model in our method, and hence represents performance without the proposed novel components.
  
For a fair comparison, we use six different prompts when evaluating GPT-4 and BioMistral:
(1) the original prompt without additional information; (2) with patient information; (3) indicating the summary is for readmission prediction; (4) for phenotype classification; (5) for both readmission prediction and phenotype classification; and (6) including all previous notes as temporal information. The original prompt is shown in the main results due to negligible performance differences between prompts (1)-(5), and poorer performance with the prompt (6). 
A single summary example is included in the prompt to ensure that the generated summaries align with structural requirements.
The content of prompts and full results for five different prompts are in \hyperlink{sec:prompt}{C.5}.
Additional results with few-shot fine-tuning settings and extractive methods are in \hyperlink{sec:fs_and_extract}{C.4}, and more details about baselines in \hyperlink{sec:baseline}{B.1}.

\subsection{Evaluation Metrics}
Evaluating the quality of text summarization is challenging \citep{bhandari2020re}, especially in specialized fields and without reference summaries. 
Therefore, we employ multiple metrics to provide a comprehensive evaluation.

\paragraph{Automatic Evaluation Metrics.}
Metrics in the automatic evaluation are divided into three categories: 1) \textit{\textbf{factuality and consistency}}, 2) \textit{\textbf{predictiveness}}, and 3) \textit{\textbf{conciseness}}.

For consistency and factuality, we consider:
(1) \textbf{UMLS-Recall}. We use QuickUMLS \citep{soldaini2016quickumls} to extract Unified Medical Language System (UMLS) biomedical concepts from the nursing note and its summary. Recall is the proportion of concepts in the original note that are present in the summary.
(2) \textbf{UMLS-FDR}. Denotes False Discovery Rate, quantifying the proportion of medical concepts in the summary that are not present in the original note. 
(3) \textbf{FactKB}. Evaluates factual consistency of summaries based on overall semantic information~\citep{feng2023factkb}.

\textbf{Predictiveness} metric assesses whether the summary adequately contains patient key information, quantified as the ability to predict the patient's condition using the summary. Specifically, we conduct readmission prediction and phenotype classification using summaries from baselines and QGSumm. For a fair comparison, we use summaries from each method to fine-tune its respective classifier for the down-stream task (similar to the query responder). This ensures the best evaluation result on predictiveness for each method, as illustrated in \hyperlink{sec:predictiveness}{B.3}. For readmission prediction, we report the weighted F1 and F1 of the positive class (``being readmitted"), and for phenotype classification, we report the F1-Macro.

Finally, we report the summary length as a percentage of the original note's length to assess conciseness.
We do not enforce a strict maximum length for baselines as the model should be capable of determining the appropriate length autonomously. 

\paragraph{Metrics used in the manual evaluation by a clinician.}
Without a reference summary, automatic evaluation metrics may not fully capture the quality of the summary.
Therefore, we invite a clinician to manually evaluate the summaries generated from 25 nursing notes by 4 summarization method (25$\times$4 summaries in total). The summaries are shown to the clinician in a randomized order and without showing the method that created them. Each summary is rated on four aspects from 1 to 5: 
(1) \textbf{Informativeness:} Whether the summary adequately captures essential information regarding the patient’s condition; 
(2) \textbf{Fluency:} Whether the summary is well-written and easy to understand.
(3) \textbf{Consistency:} How well the summary aligns with the nursing note in factuality.
(4) \textbf{Relevance:} Whether the summary is concise and not contain unnecessary information.
Detailed grading criteria are in \hyperlink{sec:grading}{B.2}.

\begin{table*}[t]
\begin{adjustwidth}{-1cm}{}
\centering
\small
\floatconts
{tab:result1}
{\caption{Results of automatic evaluation. Lower values are better for Length and UMLS-FDR, higher values for the other metrics. The subscripts denote standard deviation. ``Orig. Notes'' means using original nursing notes as such for readmission and phenotype prediction. 
``Re+Ph'' means using ``Readmission Prediction and Phenotype Classification" as the query. Results from \textbf{best} and \underline{2nd best} method under each metric are bolded and underlined. %Extractive methods are for reference and not considered in comparison.
}}
{\scalebox{0.8}{
\begin{tabular}{llcllcclc} 
\toprule
\multirow{3}{*}{Type} & \multirow{3}{*}{Method}& \multicolumn{3}{c}{Consistency and Factuality} & \multicolumn{1}{c}{Conciseness}&\multicolumn{3}{c}{Predictiveness} \\
\cline{3-9}
       &    & \multirow{2}{*}{UMLS-Recall}  & \multirow{2}{*}{UMLS-FDR} & \multirow{2}{*}{FactKB} & \multirow{2}{*}{Length} & \multicolumn{2}{c}{Readmission}      & \multicolumn{1}{c}{Phenotype~}  \\ 
\cline{7-9}
                                   &                            &  &      &   &    &\multicolumn{1}{c}{Weighted F1} & \multicolumn{1}{c}{F1}&        \multicolumn{1}{c}{Macro F1}                          \\ 
\hline
                                    & Orig. Notes              & \multicolumn{1}{c}{-}     &     \multicolumn{1}{c}{-}    & \multicolumn{1}{c}{-} & \multicolumn{1}{c}{-}  & 85.2$_{0.5}$                  &  19.7$_{1.9}$  & 28.7$_{0.5}$       \\ 
 \hline
\multirow{3}{*}{Baselines}         & BART                                                                       & \multicolumn{1}{c}{36.4$_{9.0}$}             &   \multicolumn{1}{c}{\textbf{8.70}$_{6.2}$}                                             &   \multicolumn{1}{c}{0.78$_{0.16}$}     & \multicolumn{1}{c}{\textbf{31.9\%}} & 78.8$_{0.4}$                  &  11.1$_{0.9}$  & 20.5$_{0.3}$ \\
                                   & GPT-4                                       & \multicolumn{1}{c}{\textbf{59.2}$_{8.3}$}   &      \multicolumn{1}{c}{44.2$_{7.6}$}                               &    \multicolumn{1}{c}{0.77$_{0.17}$}    & \multicolumn{1}{c}{53.6\%}  & \textbf{85.6}$_{0.6}$                  &  \textbf{21.5}$_{2.0}$  & \underline{23.6}$_{0.6}$                \\
                                   & BioMistral                             &    \multicolumn{1}{c}{55.4$_{9.9}$}     &\multicolumn{1}{c}{50.0$_{8.7}$}                         &                          \multicolumn{1}{c}{0.68$_{0.14}$}    & \multicolumn{1}{c}{69.2\%} & 80.1$_{0.6}$                  &  10.7$_{1.3}$  & 21.4$_{0.4}$  \\ 
                                   &Pegasus  &  32.2$_{8.7}$ &  \multicolumn{1}{c}{47.4$_{7.9}$}
                                 &0.65$_{0.12}$
                                   & 51.9\% & 77.1$_{0.4}$ & 9.3$_{0.8}$ & 17.5$_{0.5}$\\
\hline
% \multirow{3}{*}{10-Shot}           & BART-fs                    & 82.2$_{0.5}$                  & 14.4$_{1.3}$   & 21.1$_{0.4}$                                                    &    \multicolumn{1}{c}{52.5$_{7.3}$}          &  \multicolumn{1}{c}{44.5$_{7.1}$}                   &    \multicolumn{1}{c}{-2.72$_{0.36}$}   &    \multicolumn{1}{c}{0.76$_{0.15}$}                      &  \multicolumn{1}{c}{65.0\%}  \\
%                                    & BioMistral-fs              & 81.7$_{0.4}$                  &  10.2$_{1.1}$  & 22.0$_{0.4}$                 &       \multicolumn{1}{c}{57.2$_{10.2}$}           &\multicolumn{1}{c}{49.1$_{7.8}$}             &     \multicolumn{1}{c}{-2.97$_{0.43}$}                           &    \multicolumn{1}{c}{0.70$_{0.15}$}   &  \multicolumn{1}{c}{68.8\%}  \\
%                                    & Pegasus                    & 80.5$_{0.8}$                  &  12.5$_{1.8}$  & 18.3$_{0.6}$                 &  \multicolumn{1}{c}{35.1$_{8.4}$}             &\multicolumn{1}{c}{52.6$_{7.7}$}                 &    \multicolumn{1}{c}{-3.07$_{0.40}$}                      &    \multicolumn{1}{c}{0.70$_{0.18}$}     &  \multicolumn{1}{c}{57.4\%} \\ 
% \hline
\multirow{2}{*}{QGSumm}               & -Similarity                                                         &  \multicolumn{1}{c}{53.1$_{7.2}$}       &   \multicolumn{1}{c}{\underline{20.7}$_{6.7}$}                                                   &    \multicolumn{1}{c}{\textbf{0.82}$_{0.13}$}     &   \multicolumn{1}{c}{51.7\%} & 79.5$_{0.6}$                  & 12.0$_{1.2}$  & 22.4$_{0.4}$ \\
                                   % & -NextNote              & 80.8$_{0.6}$                  & 11.7$_{1.4}$   & 23.2$_{0.6}$               &    \multicolumn{1}{c}{56.4$_{8.0}$}          &    \multicolumn{1}{c}{35.2$_{7.1}$}                   &  \multicolumn{1}{c}{-2.32$_{0.33}$}                     &    \multicolumn{1}{c}{0.77$_{0.11}$}     &   \multicolumn{1}{c}{49.3\%}    \\
                                   % & -Readmission           & 82.4$_{0.5}$                  & \underline{18.2}$_{1.6}$   & 23.9$_{0.5}$               &   \multicolumn{1}{c}{58.2$_{7.5}$}   &     \multicolumn{1}{c}{22.7$_{6.5}$}   &     \multicolumn{1}{c} {-2.30$_{0.37}$}                      &   \multicolumn{1}{c} {0.78$_{0.14}$}     &   \multicolumn{1}{c}{\underline{46.2\%}}                          \\
                                   % & -Phenotype             & 81.9$_{0.5}$                  & 13.4$_{1.5}$  & \textbf{25.6}$_{0.6}$               &    \multicolumn{1}{c}{58.5$_{7.4}$}   &  \multicolumn{1}{c}{36.2$_{6.9}$}                 &     \multicolumn{1}{c} {-2.34$_{0.35}$}                              &   \multicolumn{1}{c} {0.79$_{0.13}$}     &  \multicolumn{1}{c}{48.0\%}  \\
                                   & -Re+Ph                &      \multicolumn{1}{c}{\underline{58.8}$_{7.9}$}   &   \multicolumn{1}{c}{24.1$_{6.4}$}                        &   \multicolumn{1}{c} {\underline{0.80}$_{0.14}$}     &  \multicolumn{1}{c}{\underline{48.2\%}}                & \underline{84.2}$_{0.5}$                  & \underline{17.2}$_{1.6}$    & \textbf{25.1}$_{0.5}$ \\ 
% \hline
% \multirow{2}{*}{Extractive} & Lead-40\%                  &    83.1$_{0.6}$                &  12.6$_{1.5}$  & 21.7$_{0.5}$                                            & \multicolumn{1}{c}{42.7$_{6.7}$}   &    \multicolumn{1}{c}{0.30$_{2.6}$}  &          \multicolumn{1}{c}{-0.87$_{0.11}$}                         &   \multicolumn{1}{c} {0.99$_{0.06}$}    &     \multicolumn{1}{c}{40.0\%}            \\
%                                    & TextRank                   &    81.9$_{0.7}$                & 14.4$_{1.7}$   & 23.3$_{0.5}$               &  \multicolumn{1}{c}{58.5$_{7.9}$}   &     \multicolumn{1}{c}{0.08$_{1.4}$}    &    \multicolumn{1}{c} {-0.90$_{0.15}$}                                   &  \multicolumn{1}{c} {0.95$_{0.12}$}      &    \multicolumn{1}{c}{51.9\%}          \\
\bottomrule

\end{tabular}
}}
\end{adjustwidth}
\end{table*}

\begin{figure*}[t]
   % \centering 
   \floatconts
   {fig:avg_score}
{\caption{Results of the manual evaluation by a clinician. The average scores in four metrics are reported for QGSumm, BART, GPT-4, and BioMistral. ``*" denotes the result of the significance test, calculated using a two-tailed Binomial test on the pairwise win-rates, i.e., we count the number of notes where QGSumm has a score higher or lower than a comparison method and test for the null hypothesis that the win-rate is 0.5.}}
    {\includegraphics[width= 1.0\linewidth]{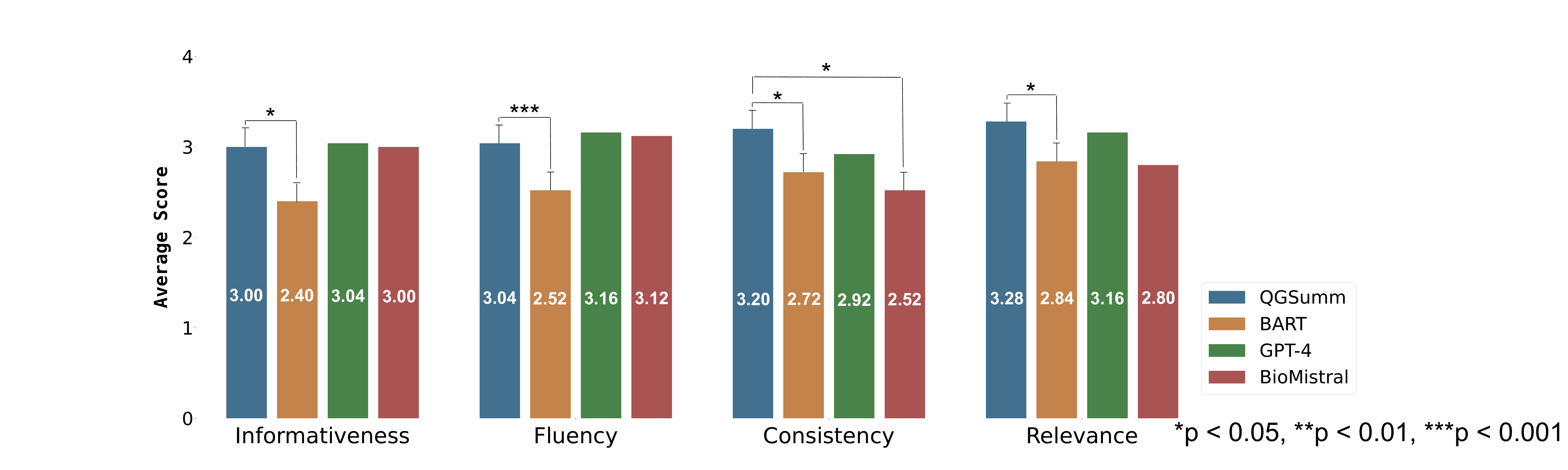}}
    % \label{fig:avg_score}
 \end{figure*}

\section{Results and Discussion}

\subsection{Automatic evaluation}

\paragraph{Conciseness, Consistency, and Factuality. }
There is a trade-off between UMLS-Recall and summary length (conciseness). As shown in Table \ref{tab:result1}, our method balance medical information consistency (measured by UMLS-Recall) and conciseness. GPT-4 captures more medical information but is less concise, whereas BART produces concise but less informative summaries. 
Similar to GPT-4, BioMistral tends to produce summaries that are not concise. Summaries from BART maintain high levels in factuality (UMLS-FDR and FactKB). Our method also has strong performance on relevant metrics.
We find that although LLMs, such as GPT-4 and BioMistral, excel in language understanding, they do not perform well on factuality. One possible reason is their tendency to rephrase or expand the original notes, potentially introducing inconsistent information, i.e. hallucination, which can cause a relatively high UMLS-FDR.

\paragraph{Predictiveness.}
Results are shown in Table \ref{tab:result1}.
In the readmission prediction task, our method closely follows GPT-4, surpassing all other baselines. We see that our method outperforms BART significantly in weighted F1 score (84.2 vs. 78.8) and F1 score of the positive class (17.2 vs. 11.1). This shows the effectiveness of the adaptation strategy guiding the model with useful queries.
Interestingly, we find that using the summary from GPT-4 for this task outperforms using the original notes, and our method's summaries perform nearly as well, highlighting their quality.
In phenotype classification, our method performs the best, outperforming BART in Macro F1 (25.1 vs. 20.5). 
Even when using similarity alone, it surpasses BART (22.4 vs. 20.5).
Notably, our method's training objective does not involve true labels of the queries, and the query responder remains frozen during training, preventing query data leakage into $\operatorname{M}$. Instead, the strong predictiveness of our summaries follows from emphasizing relevant patient-related information present in the original notes. Although specialized in text summarization, Pegasus has weak performance on all predictiveness metrics.
 
\subsection{Manual Evaluation by a Clinician}
To avoid excessive manual work, we select 3 baselines: BART, GPT-4, and BioMistral in addition to QGSumm. BART was selected as the main baseline, as it is the base model in our method but without the novel components. 
GPT-4 and BioMistral are selected due to their strong performance in the automatic evaluation.
Average scores for each method across four metrics are shown in Figure \ref{fig:avg_score}.
A case study comparing example summaries from these methods can be found in \hyperlink{sec:casestudy}{C.1}.
\paragraph{QGSumm vs BART.} 
QGSumm significantly outperforms BART on all four metrics. 
This indicates that our domain adaptation strategy enables the model to generate higher-quality summaries from the medical personnel's perspective, containing refined and important patient information with fewer hallucinations and increased readability.
Despite producing longer summaries, our method achieves a higher relevance score from the clinician, suggesting the base model struggles to identify key information and focuses on unnecessary details. Our model can effectively enhance this aspect.

\paragraph{QGSumm vs GPT-4 and BioMistral.}
% Due to the LLMs' strong language understanding capability and sufficient medical knowledge, 
% GPT-4 and BioMistral-fs can adequately summarize key information in nursing notes, receiving approximately the same average score in Informativeness as QGSumm. Additionally, they excel in generating fluent summaries by rephrasing and clarifying abbreviations, receiving a slightly higher Fluency score.
GPT-4 and BioMistral perform similarly to QGSumm in Informativeness but excel slightly in Fluency by rephrasing and clarifying abbreviations.
However, the rephrasing can introduce factual inconsistencies, and the tendency to infer additional content may reduce factuality. An example of this is provided in the case study in \hyperlink{sec:casestudy}{C.1}.
Consequently, QGSumm scores higher in Consistency, which is essential in the clinical setting. Furthermore, it generates more concise summaries with higher scores in Relevance. However, due to the small sample size, the only statistically significant difference in these comparisons was the improvement of QGSumm compared to Biomistral in consistency, and further work is needed for more conclusive results.

%\paragraph{Calculating the Significance.}
%We calculated the significance in Figure \ref{fig:avg_score} using a two-tailed Binomial test on the pairwise win-rates. In detail, we count the number of nursing notes where QGSumm has a score higher or lower than a comparison method and test for the null hypothesis that the win-probability is 0.5. THIS IS NOW IN THE CAPTION OF THE FIGURE TO SAVE SPACE!

\begin{table*}[t]
\floatconts
{tab:ablation_study}
{\caption{Results of the ablation study. We show the change in the value of the metric after removing different augmentation blocks/using different queries. $\downarrow$ denotes a decrease in the score and $\uparrow$ denotes an increase. }}
{\scalebox{0.7}{%
\begin{tabular}{cccccc} 
\toprule
            & Weighted F1 & Macro F1 & UMLS-Recall & UMLS-FDR & FactKB  \\ 
\hline
QGSumm(-Re+Ph)      & 84.2        & 25.1     & 58.8        & 24.1     & 0.80    \\ 
\hline
w/o PIA     & $\downarrow 2.6$           & $\downarrow 1.4$         &   $\downarrow 1.4$          & $\uparrow 4.7$         &  $\downarrow 0.04$       \\
w/o TIF     &  $\downarrow 4.1$            &$\downarrow 1.6$         & $\downarrow 3.8$            &$\uparrow 1.9$          &  $\downarrow 0.01$       \\
w/o PIA+TIF &   $\downarrow 4.8$           &$\downarrow 2.3$          & $\downarrow 4.4$            & $\uparrow 5.6$         &   $\downarrow 0.04$      \\
\hline
-Readmission & $\downarrow1.8$        & $\downarrow1.2$     & $\downarrow0.6$       & $\downarrow1.4$     & $\downarrow0.02$    \\
-Phenotype & $\downarrow2.3$        & $\uparrow0.5$     & $\downarrow0.3$        & $\uparrow12$     & $\downarrow0.01$    \\
\bottomrule
\end{tabular}
}}

\end{table*}

\subsection{Ablation Study}
We conducted an ablation study to: \textbf{(1)} evaluate the performance of the model without the proposed augmentation blocks: without the Patient Information Augmentation (w/o PIA), without the Temporal Information Fusion (w/o TIF), and without both blocks (w/o PIA+TIF); and \textbf{(2)} evaluate the effectiveness of the query guidance by using ``Readmission Prediction" and ``Phenotype Classification" as separate queries, instead of the combined query used earlier.

Table~\ref{tab:ablation_study} shows that the removal of augmentation blocks causes weighted F1 and macro F1 scores decrease in all settings, indicating that both blocks enhance the predictiveness of the summary. Removing TIF results in a larger score drop in F1 scores, underlining the importance of temporal information to understand the patient's current status. Conversely, the removal of PIA degrades the performance on factuality (UMLS-FDR and FactKB) more than the removal of TIF, while the removal of TIF affects more the UMLS-Recall, i.e., fewer medical concepts are captured.
On the other hand, employing different queries allows the model to focus on different aspects of the original note, resulting in summaries varying across different metrics. While there are some exceptions, the general trend is that the combined query yields overall better performance than using either query alone. Full results and a more detailed analysis are presented in \hyperlink{sec:diff_q}{C.6}.

\subsection{Discussion and Conclusion}
\label{sec:discussion}
\paragraph{User need -oriented summarization.}
A high-quality summary should facilitate efficient understanding of the relevant content by clinical personnel, especially for a nursing note the summary should capture the patient's condition. 
Our method employs patient-related queries, indirectly ensuring that the summary centers around the patient's status.
The summaries generated with different queries can be seen as coming from distinct conditional distributions and parts of the semantic space, allowing control over content and granularity.
This facilitates a more flexible and user need -oriented summary generation.
We show in \hyperlink{sec:diff_q}{C.6} that the queries can guide summaries to focus on specific aspects of the original note. For instance, broad queries produce comprehensive summaries, while detailed queries yield focused summaries on specific conditions like cardiovascular health. 

% Discussions on other topics can be found in~\hyperlink{sec:discussion_cont}{D}.

\paragraph{Design choices for information augmentation.}
One challenge is how to efficiently integrate information into the model without excessive computational cost.
We use cross-attention to allow the patient's metadata to efficiently interact on multiple levels during summary generation.
In contrast, for temporal information in previous notes, using cross-attention in a similar manner might make it difficult for the model to balance attention across the current note, past notes, and patient information, and increase computational challenges with long sequences. Hence, we represent the temporal information, obtained by weighted mean pooling from previous note representations, as the first token of the decoder's input. This strategy is intuitive, as information from previous notes naturally precedes the summary of the current note.

\paragraph{Interpretation of the evaluation metrics.}
The metrics used in the automatic evaluation have limitations as they do not conclusively reflect the quality of the summary, and come with trade-offs.  
% Also, their results lack strong correlations with those in manual evaluation.
For example, a good performance in predictiveness and medical information consistency (UMLS-Recall) may not be due to the high quality of the summary but rather caused by copying the source note, resulting in a lack of conciseness and fluency.
Conversely, as the summary becomes more concise, it may become less informative.
% Furthermore, models used to measure factuality and general consistency have inherent biases. As they are based on general semantics, they are potentially weak at recognizing patient-related information due to the dissimilarity between their training domain and clinical data, and they often prioritize text style and structure.
Furthermore, models used to measure factuality have inherent biases. They are potentially weak at recognizing patient-related information due to the dissimilarity between their training domain and clinical data.
We attempt to mitigate the impact of these limitations by comprehensively considering multiple metrics, and including the manual evaluation by a clinician.
However, there remains a need for more conclusive evaluation metrics, particularly when ground truth is unavailable.

\paragraph{Limitations.}
(1) Our current approach produces summaries of individual nursing notes, and lack the long context and support for multiple note summarization.
(2) There is room for more exploration on the formulation of the clinical queries%.
, such as generative queries.
% We don't employ generative queries but only queries related to the classification of the patient's status. 
Also, when investigating the combined effects of multiple queries, further exploration using multi-task learning methods could be beneficial.
(3) Due to the workload, the number of summaries assessed in the manual evaluation is limited to 100. 
% A larger sample size would allow more conclusive comparisons of the strengths and weaknesses of the methods.

\paragraph{Conclusion.}
We presented a novel self-supervised nursing note summarization method, where the main innovation was the introduction of query guidance, which successfully directed the summaries to include desired content. In the manual evaluation by a professional clinician, our method significantly outperformed a specialized open text summarization model, BART-Large-CNN, in all metrics. Of the other baselines, the proprietary GPT-4 had the closest performance to our method and was better than the other baselines.
In the automatic evaluation, our method outperformed GPT-4 in factual consistency, having fewer hallucinated facts without sacrificing the correct content. The same trend was seen in the manual evaluation as higher average consistency for our method.
%, although the difference was not statistically significant. 
Hence, our approach can produce more reliable summaries, clearing obstacles for responsible clinical use of LLMs. Our method demonstrates the feasibility of domain adaptation for pre-trained text summarization models without explicit supervision, and the effectiveness of self-supervised strategies to guide conditional summarization to specific interests.

\acks{This work was supported by the Research Council of Finland (Flagship programme: Finnish Center for Artificial Intelligence FCAI, and grants 352986, 358246) and EU (H2020 grant 101016775 and NextGenerationEU).}

\bibliography{reference}

\appendix

\section{Additional Details for Implementation}

\begin{figure}[t]
   \centering 
   \includegraphics[width=0.6\linewidth]{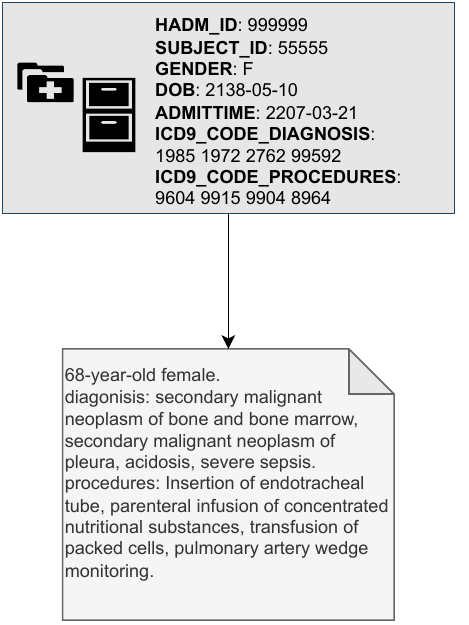} 
   \caption{Convert artificial patient metadata to a natural language description.}
   \label{fig:metadata} 
 \end{figure}

\hypertarget{sec:metadata}{\subsection*{A.1 Patient Metadata}}
Figure~\ref{fig:metadata} shows one artificial example of how patient information is obtained in natural language from structural metadata.
In the MIMIC-III database, we retrieve patient identifiers (``SUBJECT\_ID"), gender information(``GENDER"), and date of birth (``DOB") from the ``PATIENTS" table.
Information regarding admission identifiers (``HADM\_ID") and admission time (``ADMITTIME") are obtained from the ``ADMISSIONS" table, while diagnosis codes and procedure codes are sourced from ``DIAGNOSES\_ICD" and ``PROCEDURES\_ICD" tables, respectively.

\hypertarget{sec:preprocessing}{\subsection*{A.2 Data Preprocessing}}

Following prior research \citep{harutyunyan2019multitask, huang2019clinicalbert}, we perform filtering on admission records and nursing notes. Initially, we filter out specific admission cases: (1) cases of in-hospital mortality and admissions categorized as "NEWBORN"; (2) cases containing diagnosis codes outside HCUP CCS code groups. We retain only admissions containing clinical notes categorized as "Nursing/other". 
After the initial filtering, we get the statistics of The distribution of the number of nursing notes per admission, shown in Table~\ref{tab:num_notes}.
Subsequently, we apply a length limit to nursing notes, filtering out those with more than 800 tokens or fewer than 50 tokens. 
Finally, we filtered out admission cases with more than 100 nursing notes. 
Nursing notes in these cases typically represent out-of-distribution information or are irrelevant to the care episode.

We preprocess nursing notes following~\citep{huang2019clinicalbert}. 
In addition, we expand certain frequently occurring abbreviations found multiple times in each note, such as ``pt" (patient), ``cv" (cardiovascular), and ``resp" (respiratory), to aid the model's understanding of the notes.
By random sampling, we collect 10001 nursing notes from 1000 admissions as the validation set.
For the test set, we randomly select 1516 admissions and sample 3079 nursing notes from these admissions.
We only use 3079 notes for testing due to the cost of the use of GPT-4.
The nursing notes in remaining admissions are included in the training set.

\begin{table}[t]
\floatconts
{tab:num_notes}
{\caption{The distribution of the number of nursing notes per admission.}}
{{%
\scalebox{0.8}{\begin{tabular}{cc} 
\toprule
    percentiles         & number of nursing notes per admission   \\ 
\hline
  25\% & 3 \\
  50\% & 6 \\
  75\% & 13\\
  90\% & 35 \\
  100\% & 913\\
\bottomrule
\end{tabular} }
}}
\end{table}

\hypertarget{sec:uq}{\subsection*{A.3 Details of Implementation for the Query Responder}}
In addition to Readmission Prediction and Phenotype Classification, we also employ another query, Contrastive Next Note Prediction. Hence, totally five different queries are used in the experiment.

\paragraph{Contrastive Next Note Prediction.} 
Given a nursing note pair $(N, N')$, we regard the query about whether $N'$ is the successor note of $N$ as a prediction of the \textbf{patient's future status}.
To train the query responder R for the next note prediction, we create two note pairs for each nursing note, where the positive pair $(N,N_{pos})$ comprises the note and its successor in the sequence, and the negative pair $(N, N_{neg})$ contains the note and a randomly chosen non-consecutive note.
If $N$ is the patient's last nursing note, we use the patient's discharge summary and a random note from other patients to construct the positive and negative pairs.
The query is formulated as binary classification, and the output of $\operatorname{R}$ is the probability of each pair being the positive pair containing the consecutive notes:
\begin{equation}
    p_{pos} = \operatorname{R}(N,N_{pos}),\hspace{0.3cm}p_{neg} = \operatorname{R}(N, N_{neg}),
\end{equation}
\begin{equation}
    p'_{pos} = \operatorname{R}(S,N_{pos}),\hspace{0.3cm} p'_{neg} = \operatorname{R}(S, N_{neg}).
\end{equation}
The learning objective in this case is:
\begin{equation}
\label{eq:obj}
\operatorname{min}\mathcal{L}_{CE}([p_{pos}, p_{neg}], [p'_{pos}, p'_{neg}]).   
\end{equation}

\paragraph{Readmission Prediction.} 
Readmission information is easily retrieved from the hospital's database and is closely related to the patient.
The readmission prediction query is formulated as a 2-class classification task to predict whether the patient will be readmitted within 30 days of discharge, which reflects the \textbf{patient's future condition}.

The result of the readmission prediction is in the form of $[p_{pos},p_{neg}]$, indicating the probability of ``being readmitted'' and ``not being readmitted''.
The learning objective is the same as the Equation~\ref{eq:obj}.

\paragraph{Phenotype Classification.}
Classifying a patient's diagnosis status or phenotype is a query to the \textbf{patient's current status}.
Following~\cite{harutyunyan2019multitask}, phenotype classification is formulated as a multi-label classification, where ICD-9 diagnosis codes mapped by HCUP CCS code groups\footnote{\url{https://hcup-us.ahrq.gov/toolssoftware/ccs/ccsfactsheet.jsp}} are categorized into 25 classes. 
Therefore, the responder outputs the probability distribution of the phenotype as $[p_1,p_2,...,p_{25}]$. 

The prediction process can be formulated as:
\begin{equation}
    [p_1,p_2,...,p_{25}] = \operatorname{R}(N),\hspace{0.3cm}[p'_1,p'_2,...,p'_{25}] = \operatorname{R}(S).
\end{equation}

Consequently, the learning objective is:
\begin{equation}
\operatorname{min}\mathcal{L}_{CE}([p_1,p_2,...,p_{25}], [p'_1,p'_2,...,p'_{25}]). 
\end{equation}

\paragraph{Readmission Prediction and Phenotype Classification.}
We investigate the combined utilization of two queries, readmission prediction, and phenotype classification, to see if joint guidance is more effective. After obtaining the result of readmission prediction $[p^r_1,p^r_2]$ and the result of phenotype classification $[p^c_1,\ldots,p^c_{25}]$, we integrate them by converting the results into a 50-class probability distribution.

\paragraph{Training of $\operatorname{R}$.}
We use nursing notes in the training set to train the query responder $\operatorname{R}$.
The data statistics are presented in Table~\ref{tab:data_uq_training}.

To address the class imbalance issue in the readmission prediction task, we conduct oversampling for notes in the positive class (``being readmitted") and undersampling for notes in the negative class (``not being readmitted''). 
This results in 35000 nursing notes being used for training.

\begin{table}[t]
\centering
\caption{The number of nursing notes are used for training, validation and testing.}
\label{tab:data_uq_training}
\scalebox{0.8}{%
\begin{tabular}{cccc} 
\toprule
   Query         & Training & Validation & Testing  \\ 
\hline
Next Note Prediction      &   100000    &  5000  &  17458   \\ 
% \hline
Readmission Prediction      &    35000   & 10001   &  17458   \\ 
Phenotype Classification   &    149015  &  10001  &   17458  \\

\bottomrule
\end{tabular}
}
\end{table}

\hypertarget{sec:hyperparameter}{\subsection*{A.4 Hyperparameter Setting}}
We present hyperparameter settings of QGSumm and the query responder $\operatorname{R}$ in Table~\ref{tab:hp}.
The configuration of hyperparameters for the base model's architecture keeps the same as the original configuration of BART-Large-CNN\footnote{\url{https://huggingface.co/facebook/bart-large-cnn/blob/main/config.json}}.
We set the maximum length of the summary to 500 tokens, allowing for flexibility as we aim for the model to autonomously determine the appropriate length.
We use Adam optimizer~\citep{kingma2014adam} to optimize the model.
The experiments are ran on one Tesla A100 with 80G memory and one Tesla P100 with 16G memory.
\begin{table*}[t]
\centering
\caption{Details of the hyperparameter setting.}
\label{tab:hp}
\scalebox{0.8}{%
\begin{tabular}{ccc} 
\toprule
&  Hyperparameter         & Choices  \\ 
\hline
\multirow{5}{*}{QGSumm}     &  learning rate    & \{1e-5, 2e-5, 5e-5, 2e-4, 5e-4\}  \\ 
& number of training epochs & 3\\
& number of training epochs for note reconstruction & 1\\
&   $\lambda_1$   &  \{0.1, 0.3, 0.5, 0.7, 0.9, 1.0\} \\
&   $\lambda_2$   & \{0.0, 0.1, 0.3, 0.6, 0.8, 1.0\}  \\
& decoder layers being augmented by PIA & \{all 12 layers, first 6 layers, last 6 layers\} \\
\hline
\multirow{4}{*}{$\operatorname{R}$}  &learning rate & \{2e-5, 5e-5, 2e-4, 5e-4, 1e-3\}\\
& number of training epochs for next note prediction & 2 \\
& number of training epochs for readmission prediction & 2 \\
& number of training epochs for phenotype classification & 3 \\
\bottomrule
\end{tabular}
}
\end{table*}

\hypertarget{sec:next_token}{\subsection*{A.5 Next Token Prediction}}
As illustrated in \ref{sec:augmentation_block}, we obtain the hidden representation $\mathbf{H}^{dec}$ from the decoder's final layer.
The process of generating the next token then can be abstracted as:
\begin{align}
&[\mathbf{H}^{enc},\mathbf{H}^{PA}] = \operatorname{ENC}(N_i,PA),\\
&\mathbf{H}^{dec} = \operatorname{DEC}(\mathbf{H}^{enc}, \mathbf{H}^{PA}, [[TI],[BOS],y_1,\ldots,y_j]), \\
&\mathbf{v} = \operatorname{LMH}(\mathbf{H}^{dec}),\\
&\mathbf{v'} = \operatorname{ST-Gumbel Softmax}(\mathbf{v}).
\end{align}

$\operatorname{LMH}$ (Language Modeling Head) maps $\mathbf{H}^{dec}$ to a probability vector $\mathbf{v} \in \mathbb{R}^{vs}$ over the vocabulary of size $vs$. $\mathbf{v} $ is processed using the straight-though gumbel softmax trick, denoted as $\operatorname{ST-Gumbel Softmax}$, resulting in a one-hot vector $\mathbf{v'} \in \mathbb{R}^{vs}$ providing the index of the $(j+1)$th token.

\section{Additional Details for Baselines and Evaluation}

\hypertarget{sec:baseline}{\subsection*{B.1 Baselines}}

\subsubsection*{Choice of the baselines}
\textbf{BART-Large-CNN}~\citep{lewis2020bart}: It is chosen as the base model for its excellent performance on text summarization as well as less computation cost than its peers.
We consider it as one baseline in the experiment to illustrate performance without the proposed novel components.
\textbf{Pegasus}~\citep{zhang2020pegasus}: It is a transformer-based pre-trained model specialized in abstractive summarization, widely recognized as a baseline model in many studies on text summarization.
\textbf{BioMistral-7B}~\citep{labrak2024biomistral}:
It is an open-source instruction-based LLM adapted from Mistral~\citep{jiang2023mistral} for the medical domain. 
It achieves state-of-the-art performance in supervised fine-tuning benchmarks compared to other open-source medical language models.
\textbf{GPT-4}~\citep{openai2023gpt}: It is a proprietary LLM representing state-of-the-art on general NLP task.
We employ one-shot in-context learning to prompt GPT-4 and BioMistral-7B, providing one summary example to ensure the structure of generated summaries aligns with the requirements.
We show prompt examples in \hyperlink{sec:prompt}{C.5}.
Consistent with our method's settings, the maximum length of the summary is set to 500 tokens.

In Table~\ref{tab:num_parameters}, we provide the statistics on the total number of parameters for QGSumm and baselines.

\begin{table}[t]
\floatconts
{tab:num_parameters}
{\caption{The statistics on the total number of parameters for QGSumm and baselines (except GPT-4, whose parameter count remains undisclosed).}}
{{%
\begin{tabular}{cc} 
\toprule
             & number of parameters   \\ 
\hline
  QGSumm & 457M \\
  BART-Large-CNN & 406M \\
  Pegasus & 568M\\
  BioMistral & 7B \\
\bottomrule
\end{tabular}
}}
\end{table}

We also consider two representative extractive methods, TextRank~\citep{mihalcea2004textrank} and Lead-40\%, and report the evaluation results of them as a reference. We do not compare our methond and baselines with them, since extractive methods are out of the scope of this work.
In TextRank, we utilize MPNet~\citep{song2020mpnet} to obtain sentence embeddings. 
In Lead-40\%, we use the first 40\% of the content of the note as a summary.

\subsubsection*{Few-shot Adaptation}
In addition, we show the performance of baselines in few-shot fine-tuning settings for reference.
We randomly sample 10 nursing notes from the training set and pair them with their corresponding summaries generated by GPT-4 as the sliver ground truth to create the training data for 10-shot fine-tuning of BART-Large-CNN, BioMistral-7B, and Pegasus.
The training data is transformed into instructions-formatted prompts for fine-tuning BioMistral-7B.
We fine-tune BART-Large-CNN and Pegasus for 8 and 9 epochs, respectively, with a learning rate of 0.0005.
As for BioMistral-7B, we fine-tune it using QLoRA adaptation for 7 epochs with a learning rate set to 0.0002.

\hypertarget{sec:grading}{\subsection*{B.2 Grading Criteria of Manuel Evaluation}}
\begin{figure*}[t]
   \centering 
   \includegraphics[width=0.7\linewidth]{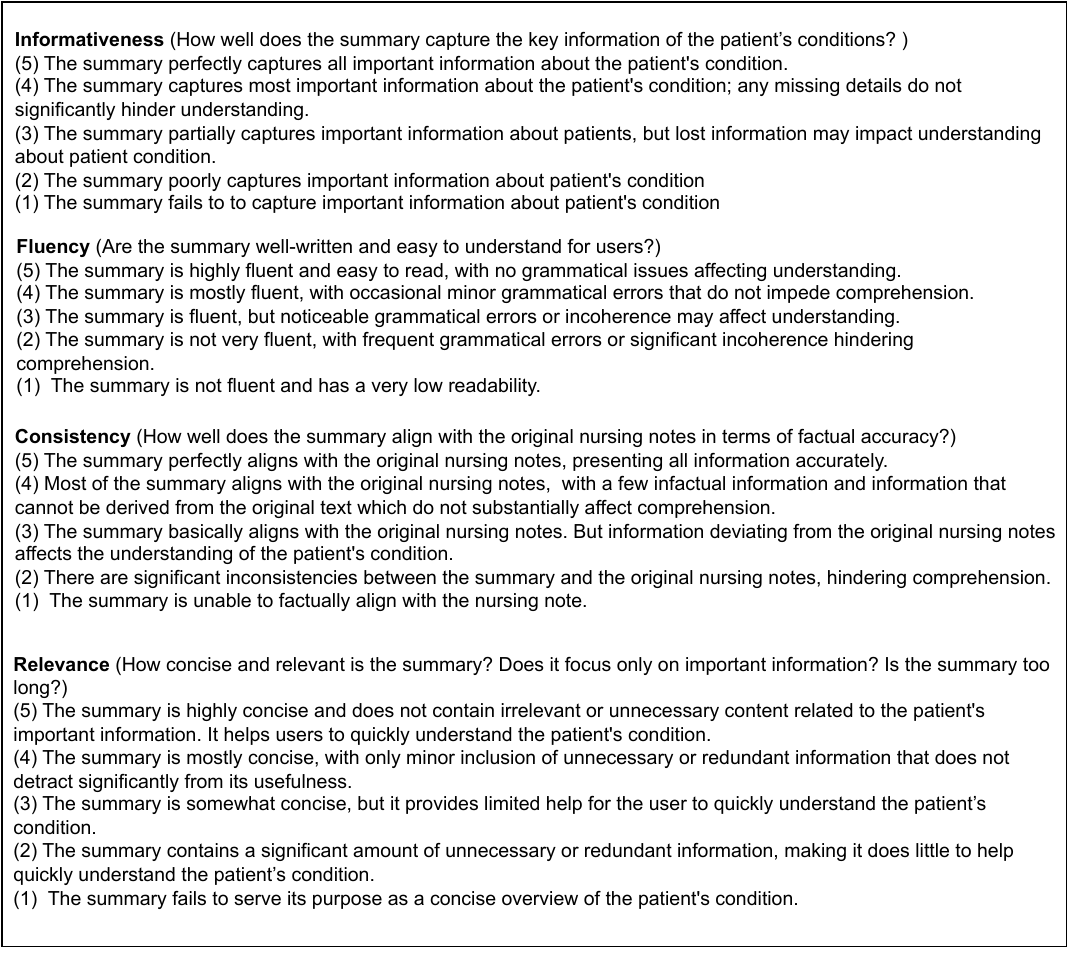} 
   \caption{The grading criteria provided to the clinician for manual evaluation.}
   \label{fig:grading} 
 \end{figure*}
Figure~\ref{fig:grading} shows the detailed grading criteria we provide to the clinician for manual evaluation. 
The score for each metric ranges from 1 to 5.

\hypertarget{sec:predictiveness}{\subsection*{B.3 Evaluations on Predictiveness}}
Based on the assumption that a high-quality summary, by effectively capturing critical patient information, can better predict both the current and future status of patients, we introduce predictiveness as a novel metric to evaluate the quality of the summaries.

Specifically, we assess predictiveness by measuring the performance of readmission prediction and phenotype classification using summaries generated by various baselines. The evaluation workflow for this metric is as follows:
\begin{itemize}
    \item Generate summaries of notes in the test set for each method (baselines and QGSumm);
    \item For summaries from one method, split them into 10 folds; 
    \item Fine-tune the \textbf{own} query responder of each method using its summaries, and compute the metrics through 10-fold cross-validation.
\end{itemize}

For instance, to evaluate the predictiveness of summaries from BART, we begin with the trained query responder $\operatorname{R}$ as described in \hyperlink{sec:uq}{A.3}. We fine-tune $\operatorname{R}$ using the BART-generated summaries of nursing notes from the test set and then compute the relevant metrics through 10-fold cross-validation. 
Consequently, we obtain multiple responders tailored to each method, specifically for readmission prediction and phenotype classification.

We fine-tune each method's query responder using its own summaries to \textbf{ensure the most accurate evaluation of predictiveness} for each method. For comparison, we also experimented with using only GPT-4's summaries to fine-tune a single query responder across all methods. As shown in Table~\ref{tab:eval_pred}, the results got worse for all methods distinct from GPT-4, which might be due to the dissimilarity of the semantic spaces between summaries from GPT-4 and summaries from others. 
By optimizing a separate predictor for each model we get the best possible predictions for each one, leading to a fair comparison.

\begin{table}[t]
\floatconts
{tab:eval_pred}
{\caption{Reults on predictiveness get worse when using only summaries from GPT-4 to finetune the query responder. Weighted F1 and Macro F1 are reported for readmission prediction and phenotype classification, respectively. }}
{{%
\begin{tabular}{ccc} 
\toprule
            & Weighted F1 & Macro F1   \\ 
\hline
  QGSumm & $\downarrow${1.6} & $\downarrow${0.7}\\
  BART & $\downarrow${1.9} & $\downarrow${0.9}\\
  BioMistral & $\downarrow${0.9} & $\downarrow${0.4}\\
  Pegasus & $\downarrow${1.7} & $\downarrow${0.7} \\
\bottomrule
\end{tabular}
}}
\end{table}

\label{sec:case}
\begin{figure*}[t]
   \centering 
   \includegraphics[width=0.7\linewidth]{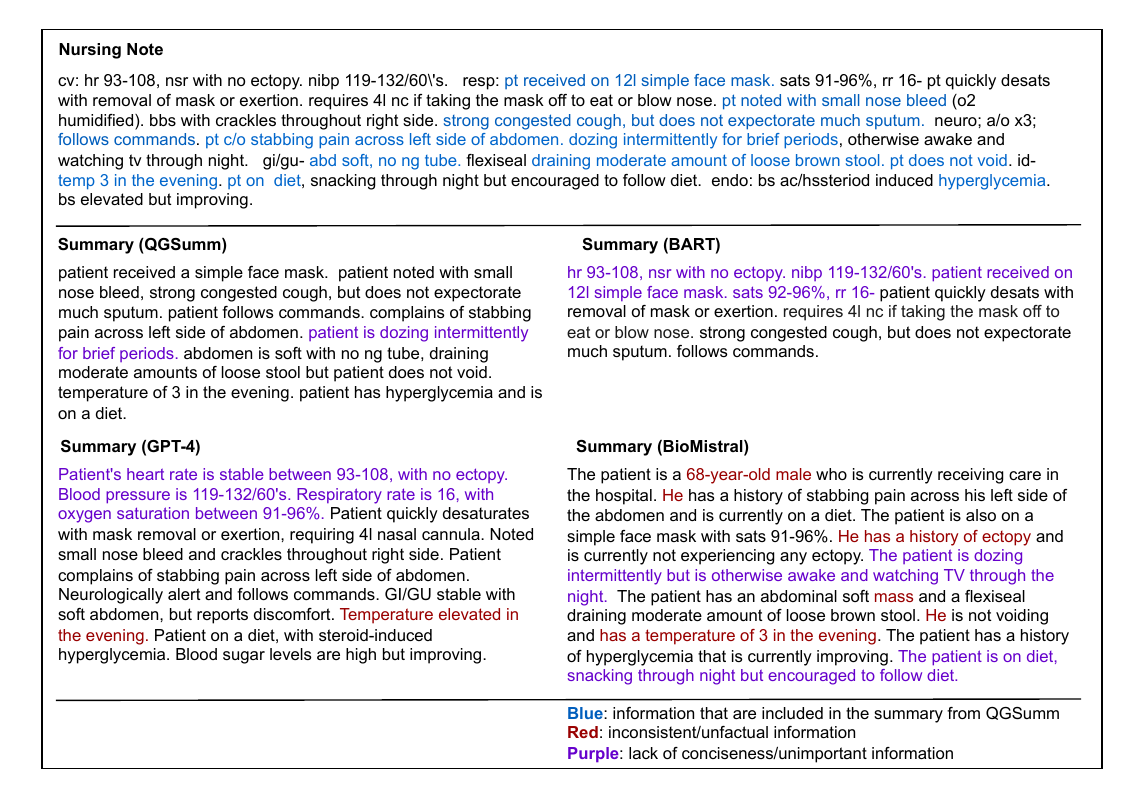} 
   \caption{One artificial nursing note and its summaries from QGSumm, BART, GPT-4, and BioMistral, respectively.}
   \label{fig:case_study} 
 \end{figure*}

 \begin{figure*}[t]
   \centering 
   \includegraphics[width=0.8\linewidth]{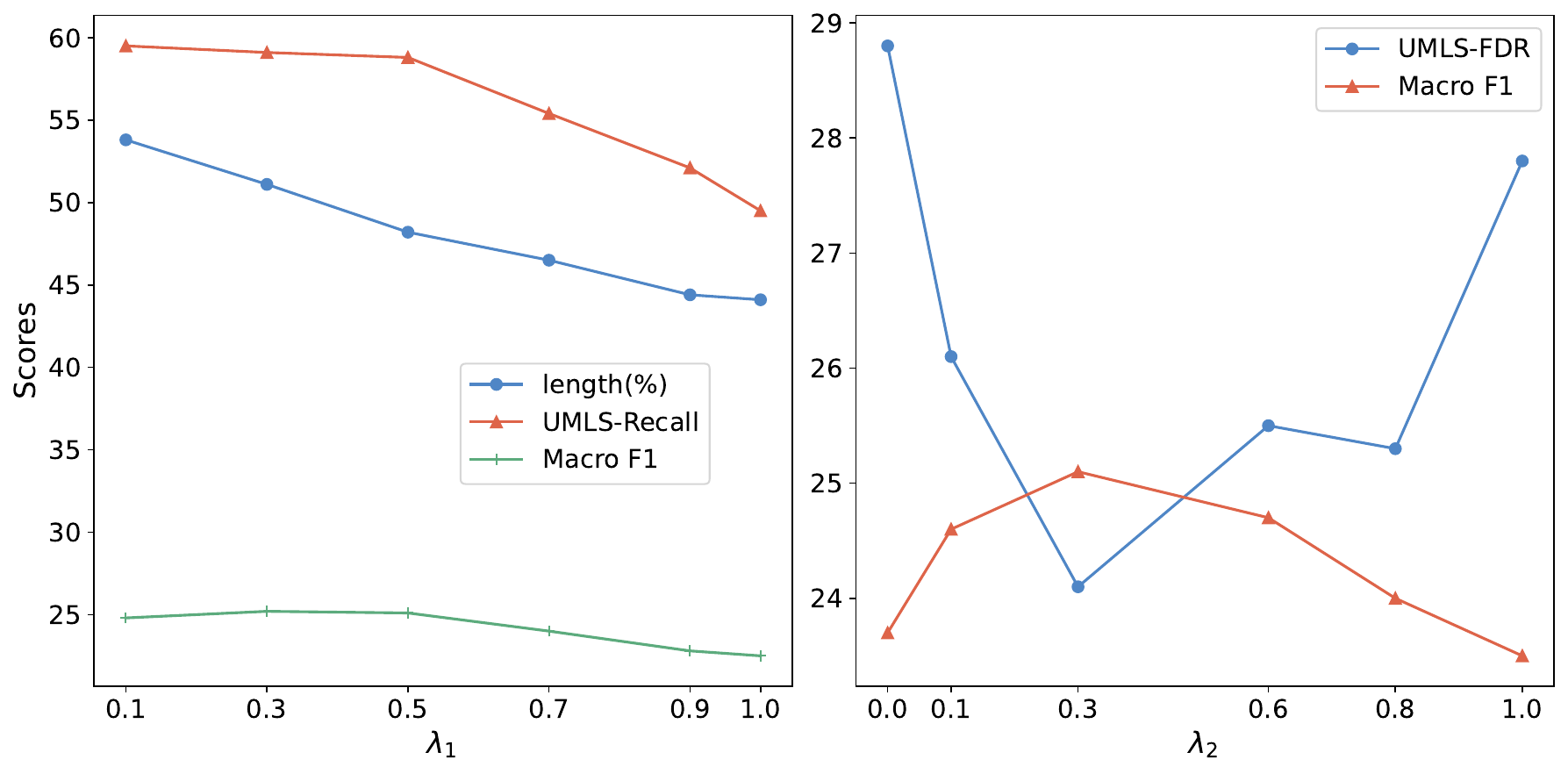} 
   \caption{Effects of various values of $\lambda_1$ (Equation~\ref{eq:training_obj}) and $\lambda_2$ (Equation~\ref{eq:PIA}). (a): $\lambda_1$ affects both the length and the medical information consistency of the generated summary, resulting in changes in the summary's predictiveness; (b): $\lambda_2$ regulates the importance of the PIA block. The PIA is applied to all decoder layers. }
   \label{fig:lambda} 
 \end{figure*}
 
\section{Additional Experimental Results}
\hypertarget{sec:casestudy}{\subsection*{C.1 Case Study}}
One artificial nursing note and its corresponding summaries generated by QGSumm, BART, GPT-4, and BioMistral are presented in Figure~\ref{fig:case_study}.
In the original nursing note, the content highlighted in blue indicates information included in the summary generated by our approach.
We can see that our approach captures most of the important patient information. 
However, some details, such as cardiovascular and respiratory conditions, are overlooked.
The summary from BART only covers information from the first half of the nursing note, suggesting the limitation in understanding long context.
Summaries from GPT-4 and BioMistral contain more patient information but lack conciseness.
These models achieve fluency by rephrasing notes and expanding abbreviations.
However, BioMistral struggles with maintaining factuality, often excessively reasoning about the patient's personal information and condition.

 \begin{table*}[t]
\centering
\small
\caption{Additional results of: (1) automatic evaluation on few-shot fine-tuning and extractive methods; (2) using five different queries. Lower values are better for Length and UMLS-FDR, higher values for the other metrics. The subscripts denote standard deviation.}
\setlength{\tabcolsep}{5pt}
\scalebox{0.8}{
\begin{tabular}{llcllcclc} 
\toprule
\multirow{3}{*}{Type} & \multirow{3}{*}{Method}& \multicolumn{3}{c}{Consistency and Factuality} & \multicolumn{1}{c}{Conciseness}&\multicolumn{3}{c}{Predictiveness} \\
\cline{3-9}
       &    & \multirow{2}{*}{UMLS-Recall}  & \multirow{2}{*}{UMLS-FDR} & \multirow{2}{*}{FactKB} & \multirow{2}{*}{Length} & \multicolumn{2}{c}{Readmission}      & \multicolumn{1}{c}{Phenotype~}  \\ 
\cline{7-9}
                                   &                            &  &      &   &    &\multicolumn{1}{c}{Weighted F1} & \multicolumn{1}{c}{F1}&        \multicolumn{1}{c}{Macro F1}                          \\ 
\hline

\multirow{3}{*}{Few-Shot}           & BART   &    \multicolumn{1}{c}{52.5$_{7.3}$}          &  \multicolumn{1}{c}{44.5$_{7.1}$}                      &    \multicolumn{1}{c}{0.76$_{0.15}$}                      &  \multicolumn{1}{c}{65.0\%}                  & 82.2$_{0.5}$                  & 14.4$_{1.3}$   & 21.1$_{0.4}$                                                     \\
 & BioMistral  &       \multicolumn{1}{c}{57.2$_{10.2}$}           &\multicolumn{1}{c}{49.1$_{7.8}$}             &        \multicolumn{1}{c}{0.70$_{0.15}$}   &  \multicolumn{1}{c}{68.8\%}            & 81.7$_{0.4}$                  &  10.2$_{1.1}$  & 22.0$_{0.4}$                   \\
& Pegasus   &  \multicolumn{1}{c}{35.1$_{8.4}$}             &\multicolumn{1}{c}{52.6$_{7.7}$}                 &    \multicolumn{1}{c}{0.70$_{0.18}$}     &  \multicolumn{1}{c}{57.4\%}                 & 80.5$_{0.8}$                  &  12.5$_{1.8}$  & 18.3$_{0.6}$                  \\ 
\hline
\multirow{2}{*}{Extractive} & Lead-40\%    & \multicolumn{1}{c}{42.7$_{6.7}$}   &    \multicolumn{1}{c}{0.30$_{2.6}$}  &            \multicolumn{1}{c} {0.99$_{0.06}$}    &     \multicolumn{1}{c}{40.0\%}                &    83.1$_{0.6}$                &  12.6$_{1.5}$  & 21.7$_{0.5}$                                                      \\
& TextRank     &  \multicolumn{1}{c}{58.5$_{7.9}$}   &     \multicolumn{1}{c}{0.08$_{1.4}$}                                      &  \multicolumn{1}{c} {0.95$_{0.12}$}      &    \multicolumn{1}{c}{51.9\%}                  &    81.9$_{0.7}$                & 14.4$_{1.7}$   & 23.3$_{0.5}$                     \\
\hline
\multirow{5}{*}{QGSumm} & -Similarity  &                        \multicolumn{1}{c}{53.1$_{7.2}$}       &   \multicolumn{1}{c}{20.7$_{6.7}$}                                                   &    \multicolumn{1}{c}{0.82$_{0.13}$}     &   \multicolumn{1}{c}{51.7\%} & 79.5$_{0.6}$                  & 12.0$_{1.2}$  & 22.4$_{0.4}$ \\
 & -NextNote  &       \multicolumn{1}{c}{56.4$_{8.0}$}   &   \multicolumn{1}{c}{35.2$_{7.1}$}                                   &    \multicolumn{1}{c}{0.77$_{0.11}$}     &   \multicolumn{1}{c}{49.3\%}             & 80.8$_{0.6}$                  & 11.7$_{1.4}$   & 23.2$_{0.6}$                           \\
 & -Readmission   &   \multicolumn{1}{c}{58.2$_{7.5}$}   &     \multicolumn{1}{c}{22.7$_{6.5}$}   &     \multicolumn{1}{c} {0.78$_{0.14}$}                       &   \multicolumn{1}{c}{46.2\%}           & 82.4$_{0.5}$                  & 18.2$_{1.6}$   & 23.9$_{0.5}$                                       \\
 & -Phenotype  &    \multicolumn{1}{c}{58.5$_{7.4}$}   &  \multicolumn{1}{c}{36.2$_{6.9}$}                                             &   \multicolumn{1}{c} {0.79$_{0.13}$}     &  \multicolumn{1}{c}{48.0\%}            & 81.9$_{0.5}$                  & 13.4$_{1.5}$  & 25.6$_{0.6}$                 \\
                                   & -Re+Ph                &      \multicolumn{1}{c}{58.8$_{7.9}$}   &   \multicolumn{1}{c}{24.1$_{6.4}$}                        &   \multicolumn{1}{c} {0.80$_{0.14}$}     &  \multicolumn{1}{c}{48.2\%}                & 84.2$_{0.5}$                  & 17.2$_{1.6}$    & 25.1$_{0.5}$ \\ 
\bottomrule
\end{tabular}
}
\label{tab:result1_fs_extractive_query}
\end{table*}
 
\hypertarget{sec:length_coef_effect}{\subsection*{C.2 Effects of the Length Penalty Coefficient}}
The effects of varying the length penalty coefficient are shown in Figure~\ref{fig:lambda}(a).
When $\lambda_1$ increases, the generated summaries become more concise.
However, once $\lambda_1$ exceeds 0.5, there is a notable decrease in medical information consistency, accompanied by a decline in predictiveness performance.
One potential explanation for this phenomenon is that within the range of 0.1 to 0.5, $\lambda_1$ facilitates the refinement of nursing notes by filtering out unnecessary information. 
However, surpassing 0.5 in the value of $\lambda_1$ results in a stricter penalty, which causes the omission of the patient key information for obtaining more concise summaries.
To strike a balance between conciseness and consistency, we ultimately set $\lambda_1$ to 0.5.

\hypertarget{sec:importance_effect}{\subsection*{C.3 Effects of the Importance of Patient Meta Information}}
$\lambda_2$ regulates the contribution of patient metadata to the summarization process.
The incorporation of patient meta information helps maintain the factuality of the summary.
However, as shown in Figure~\ref{fig:lambda}(b), this influence is not consistently beneficial, with the optimal effect observed when $\lambda_2$ is set to 0.3. Additionally, excessively large $\lambda_2$ causes the model to prioritize patient metadata over the content of the nursing note, which degrades the quality of the generated summary, reflected as reduced predictiveness.
Therefore, we set $\lambda_2$ to 0.3 for our method.

\hypertarget{sec:fs_and_extract}{\subsection*{C.4 Results on Few-shot and Extractive Methods}}
We present additional results on few-shot fine-tuning settings and extractive methods, as shown in Table~\ref{tab:result1_fs_extractive_query}.

After fine-tuning with 10 summaries generated by GPT-4, BART and Pegasus produce summaries that include more medical and patient-related information, as indicated by higher F1-score and UMLS-Recall.
However, their performance declines in certain factual metrics and their summaries become less concise, mirroring the performance trend seen with GPT-4.
BioMistral's performance has not changed much after few-shot adaptation.
Considering the inherent bias from the limitations of GPT-4, we do not compare QGSumm with few-shot fine-tuning methods in the main text. 

\begin{table*}[t]
\centering
\caption{We conduct evaluations on 100 nursing notes using six different prompts for GPT-4 and BioMistral. Lower values are better for Length and UMLS-FDR, higher values for the other metrics.}
\label{tab:prompts}
\scalebox{0.8}{%
\begin{tabular}{cccccccc} 
\toprule
& &     UMLS-Recall      & UMLS-FDR & FactKB & Length & Weighted F1 & Macro F1  \\ 
\hline
\multirow{6}{*}{GPT-4}     &    Original Prompt   & 56.1 & 44.9 & 0.73 & 56.5\% & 84.6 & 23.4 \\ 
& - Patient Information & 53.8 & 48.0 & 0.76 & 63.6\% & 83.2 & 21.7\\
& - Readmission & 55.6 & 44.4 & 0.64 & 54.5\% & 83.5 & 24.0\\
&- Penotype & 54.4 & 45.1 & 0.61 & 54.9\% & 83.5 &  21.1 \\
&- Re+Ph & 56.1 & 45.0 & 0.64 & 55.7\% & 84.6 & 22.6 \\
&- Temporal Info & 52.2 & 50.2 & 0.65 & 60.3\% & 82.4 & 21.5 \\

\hline
\multirow{6}{*}{BioMistral}     &    Original Prompt   & 53.8 & 50.3  & 0.69 & 67.7\% & 79.7 & 21.2 \\ 
& - Patient Information & 55.1& 54.4& 0.64& 70.1\% & 78.6 & 22.0\\
& - Readmission &53.1&50.3&0.63&67.1\%&78.2&20.2\\
& - Penotype &52.2&51.2 &0.63&65.9\%&77.6&20.9   \\
& - Re+Ph &53.8&51.7&0.61&66.7\%&79.3& 21.4  \\
& - Temporal Info &50.4&58.6&0.62&68.2\%&76.0& 19.3  \\
\bottomrule
\end{tabular}
}
\end{table*}

 \begin{figure*}[t]
   \centering 
   \includegraphics[width=0.9\linewidth]{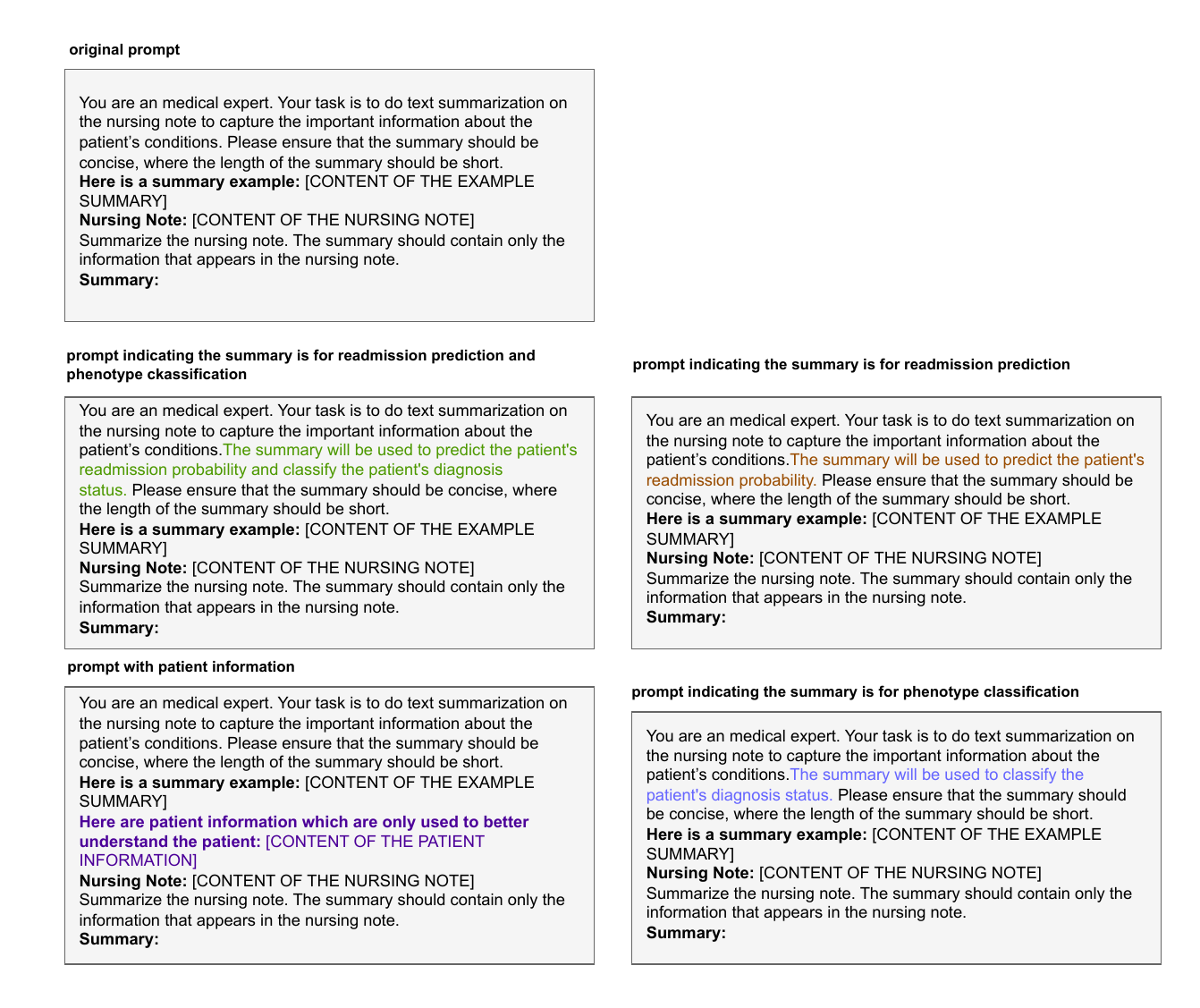} 
   \caption{We use five different prompts for evaluating GPT-4 and BioMistral.}
   \label{fig:prompts} 
 \end{figure*}

\hypertarget{sec:prompt}{\subsection*{C.5 Prompts for GPT-4 and BioMistral}}
Examples of the prompts used for evaluating GPT-4 and BioMistral are shown in Figure~\ref{fig:prompts}. We observe that when a summary example is not included in the prompt, the model sometimes generates outputs that fail to meet the structural requirements, such as producing lists of sentences. This issue is more prevalent when using GPT-4.

To determine the optimal prompt content, we conduct evaluations using six different prompts, with the results shown in Table~\ref{tab:prompts}. 
Given the cost of GPT-4, we limit the evaluation to 100 randomly selected nursing notes from the test set. Compared to the original prompt, adding patient information or queries did not result in significant differences in performance for either GPT-4 or BioMistral. 
When considering temporal information by including all previous notes in the prompt, we observed that including temporal information interfered with the summarization of the current note, causing both GPT-4 and BioMistral to miss key information from the current note.
Consequently, we choose to use the original prompts in the main experiments, as they yielded more balanced results across different metrics.

Furthermore, we note that unlike our method, using prompts that explicitly mention readmission prediction or phenotype classification does not improve performance on those specific tasks. This highlights the effectiveness of using the query to parametrically guide the model's behavior.

\hypertarget{sec:diff_q}{\subsection*{C.6 Effectiveness of the Query Guidance}}
We provide full results of employing five different queries shown in Table~\ref{tab:result1_fs_extractive_query}.
We can observe: (1) Regarding predictiveness, employing queries closely related to patients and focusing on readmission and phenotype information yield superior performance compared to other queries.
As expected, the method with phenotype-related queries performs the best in phenotype classification, while the method with readmission-related queries is the best in readmission prediction. 
This highlights the effectiveness of guiding the summarization with queries, and \textbf{different queries enable the summary to concentrate on different aspects of the original note.}
(2) Using similarity as guidance can produce summaries that are more similar to the original notes, resulting in higher scores on general consistency and factuality.
However, summaries generated under this configuration tend to be longer and often sacrifice predictiveness and informativeness regarding medical concepts, \textbf{demonstrating the limitations of the unconstrained guidance signal.}
(3) Using Next Note Prediction as the query results in weaker performance compared to patient-related queries, indicating that \textbf{queries lacking meaningful context fail to guide the model effectively.}
(4) When employing joint guidance with both readmission and phenotype information, our method consistently achieves excellent performance across all metrics.
This indicates that \textbf{combining different guidance signals can help in producing better summaries,} and further research is needed to explore this aspect in depth.

\end{document}